\documentclass[cameraready]{Interspeech}

\title{Adding Robust Code-Switching Capabilities to High Performance Multilingual ASR}
\author[affiliation={1}, orcid=0009-0009-5181-3910]{Enes Yavuz}{Ugan}
\author[affiliation={1,2}]{Alexander}{Waibel}

\address{
    $^1$ 
Interactive Systems Lab, Karlsruhe Institute of Technology (KIT), Germany \\
    $^2$ 
InterACT, Carnegie Mellon University (CMU), USA
}

\email{enes.ugan@kit.edu}


\usepackage{comment}

\usepackage{tikz}
\usepackage{pgfplots}
\pgfplotsset{compat=1.18}
\usepackage{graphicx}   
\usepackage{amsmath}    
\usepackage{xcolor}     
\usepackage{tcolorbox}
\usepackage{tipa} 
\newcommand{\huggingfacesmall}{\includegraphics[width=9px]{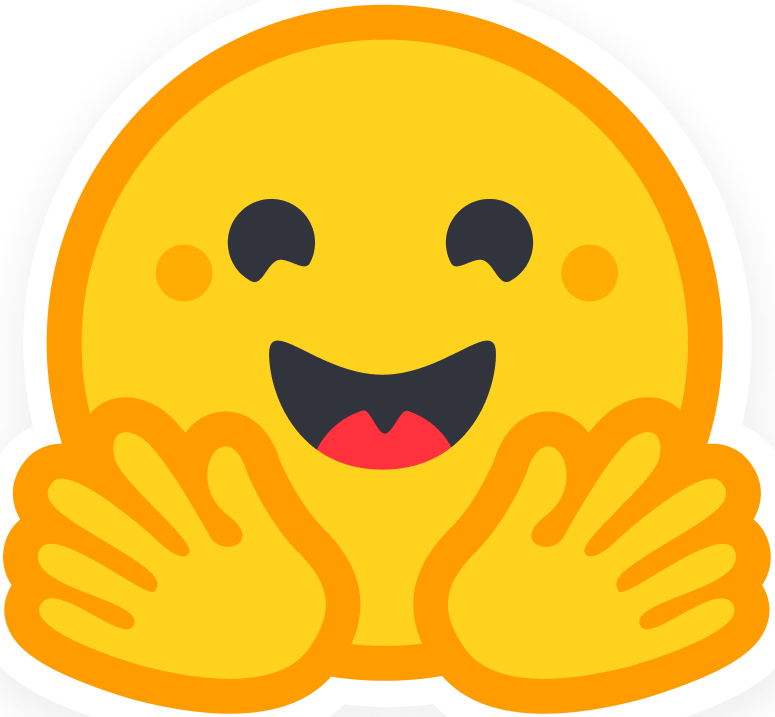}}
\newcommand{\githubsmall}{\includegraphics[width=9px]{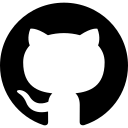}}

\begin{document}

\maketitle

\begin{abstract}
 Code-switching (CSW) remains challenging for large multilingual ASR systems in real-world deployment. While fine-tuning on synthetic CSW data is possible, it generally degrades strong monolingual baselines. Our goal is to preserve these capabilities while extending models to handle complex code-switching, including morphological variations across languages. We propose Bayesian factorized adaptation, which learns to efficiently integrate switching-relevant knowledge into strong pretrained models without overwriting existing capabilities. Requiring only a small amount of synthetic data, our approach reduces transcription errors by 32.87\% on code-switched words while improving overall WER by 5.31\%, all while maintaining monolingual performance. Our results demonstrate that effective CSW adaptation depends more on knowledge integration than data complexity.
\end{abstract}

\section{Introduction}

Code-switching automatic speech recognition addresses the growing need to recognize multilingual speech where speakers seamlessly alternate between languages. 
While multilingual speech recognition has been studied for decades through language identification \cite{schultz1996lvcsr}, cross-language acoustic modeling \cite{schultz2001experiments}, and multilingual articulatory feature integration \cite{stuker2003integrating}, CSW ASR presents additional challenges that go beyond standard multilingual recognition.
Real-world code-switching is data scarce, as annotated CSW corpora remain rare and expensive to collect. Furthermore, switching can occur at multiple linguistic levels, between sentences, at the lexical level, or even sub-lexically through morphological integration of foreign words into the matrix language. 
 For example, an English verb like "download" may appear as "downgeloadet" in German-English CSW, receiving full German morphological inflection. 
 Achieving robust CSW recognition therefore requires the model to internalize syntactic ordering constraints across languages, knowledge that cannot be acquired from surface-level data augmentation alone. Despite these challenges, most code-switching ASR research can be categorized into one of three well-established paradigms, each with significant methodological limitations. We identify a fourth, underexplored scenario that represents a practically important and challenging research direction.

\textbf{Scenario 1: In-Domain code-switching Optimization.} A substantial body of work focuses exclusively on improving code-switching performance on specialized code-switching test sets (e.g., SEAME \cite{lyu2010seame}, ArzEn \cite{hamed2020arzen}, Fisher \cite{weller2022end}, DECM \cite{ugan2024decm}) without systematic evaluation of monolingual capabilities. Representative examples include LLM-based synthetic code-switching text generation \cite{alharbi2024leveraging} and TTS-augmented training for low-resource pairs \cite{sharma2020improving} and many others, such as architectural approaches \cite{chowdhury2021towards,ali2021arabic}, or adjustment of loss functions \cite{ugan2025adapting}. 
While these approaches demonstrate in-domain WER or Mixed-Error-Rate (MER) improvements, they cannot distinguish whether gains reflect genuine code-switching robustness or overfitting to narrow acoustic/linguistic characteristics of the code-switching dataset.

\textbf{Scenario 2: In-Domain Multilingual Evaluation.} More recent work combines monolingual and synthetic code-switching data during training and evaluates on both modalities \cite{nguyen2025can, yan2025cs,ugan2022language,seki2018end}. However, monolingual test sets typically originate from the same acoustic domains and speaker populations as training data. Such evaluation does not probe robustness to \textit{out-of-domain} cases; accented speech, far-field audio, or conversational patterns. Thus, claims of dual-objective success remain unverified on truly diverse, out-of-domain monolingual speech, what we term the \textit{in-domain monolingual validation problem}.

\textbf{Scenario 3: Weak Baseline Adaptation.} A third category starts from weak ASR models (trained from scratch or pre-trained models that are generally strong but have poor language specific performance) and reports joint improvements on both monolingual and code-switching tasks \cite{zhang2022reducing, babatunde2025beyond}. While genuine, improvements are confounded by weak starting baselines, making it unclear whether methods scale to strong production-quality models where decision boundaries are near-optimal and vulnerable to distribution shift.

\textbf{Scenario 4: Strong Multilingual Model Preservation (Our Focus).} The most challenging and practically important scenario is to \textit{improve} code-switching while \textit{preserving} the capabilities of strong pre-trained multilingual models. Strong multilingual ASR (e.g., Whisper-large, XLS-R) is valued in production precisely because it reliably recognizes diverse languages and acoustic conditions. Adding code-switching support is a secondary objective that must not degrade this general capability. However, this scenario has received limited attention in the literature. Most previous work such as merging of LoRA adapters \cite{ugan2025weight}, soft-prompting \cite{yang2025adapting}, or sparse LoRA adapters \cite{ugan2026bayesian} need real code-switching training data to be useful. 

In each scenario, work without real training data mainly explores different ways of generating synthetic data and utilizing it for improving code-switching capabilities.
Work in \cite{sharma2020improving} tackles scenario 1: they introduce a feature level mixup of TTS and real data, as well as a code-switching bias loss enforcing prediction of English tokens. 
Their model was trained from scratch and evaluated on their code-switching test data, only.
In \cite{du2021data} the authors cover scenario 3, by exploring three different data generation approaches such as audio splicing of existing code-switching data, and TTS with word translations or insertions. 
The authors in \cite{nguyen2025can} also apply TTS splicing to sentence translations and aim to improve larger pre-trained speech models, focusing on scenario 2.
In this work, we contribute to this line of work (Scenario 4) by improving code-switching ASR without the need of any real code-switched data, starting from an already strong baseline.

\subsection{Motivation and Key Contribution}
A widely held assumption in code-switching ASR research is that better synthetic data leads to better adaptation \cite{nguyen2025can, yan2025cs,ugan2022language,seki2018end}. This assumption has been validated repeatedly, but almost exclusively on weaker baselines, where any additional training signal is beneficial. 
We challenge this assumption by asking:
\textit{Does synthetic data actually help when the model is already strong?}
We adapt a high-performing Whisper model on English-German; one of its strongest language pairs; using only LLM-generated code-switching text and competitive TTS synthesis, without any real code-switching speech. 
This controlled setting allows us to isolate the effect of the adaptation mechanism from the effect of data quality. 
Our main findings are:
\begin{itemize}
    \item \textbf{Synthetic data alone is not enough, and can be actively harmful.} 
    Standard LoRA fine-tuning on synthetic code-switching data causes up to 418\% relative WER degradation on German and over 200\% relative increase in PIER, even with carefully designed data pipelines. 
    This holds across data quantity and synthesis strategy, demonstrating that naive fine-tuning is fundamentally ill-suited to adapting already-strong multilingual models.
    
    \item \textbf{The bottleneck is integration, not data.} Increasing synthetic data complexity; through multi-step pipelines, translation alignment, or audio splicing; does not resolve this degradation. 
    The critical variable is how new linguistic knowledge is merged with existing model knowledge, not how the data is generated.

    \item \textbf{Principled integration turns synthetic data into a net benefit.}
    Bayesian Low-Rank Adaptation (BLoRA) \cite{ugan2026bayesian}, addresses this by learning sparse, uncertainty-aware adaptation matrices that selectively integrate switching-relevant knowledge while preserving the model's existing capabilities. 
    Combined with a simplified two-step synthesis pipeline (LLM prompting + TTS), BLoRA achieves a 32.87\% relative PIER reduction and a 5.31\% relative WER improvement on CSFleurs, while maintaining monolingual performance on CommonVoice.
    
    \item \textbf{A deployable, annotation-free pathway.} 
    Because our approach requires no real code-switching speech, no manual annotation, and no language-pair-specific components beyond basic linguistic prompting rules, it offers a scalable pathway to enabling code-switching in production systems for any language pair where no CSW corpora exist.
\end{itemize}

\section{ Experimental Design and Motivation}
\label{sec:experimentaldesign}
\subsection{Deliberate Choice of Language Pair: English-German as a Challenge}
We deliberately select English-German code-switching to evaluate our approach under the most challenging conditions: 
The Whisper model achieves WER of 8.53\% on German and 13.56\% on English (CommonVoice 14 \cite{ardila2020common}), among its strongest performances across all language pairs. This choice is deliberate: if we can improve code-switching performance on an already-strong language pair without degradation, the method's value for weaker pairs and more distant language families is even more compelling. In contrast, demonstrating improvements on weak baselines provides limited evidence of practical utility.

Our evaluation uses the most recent German-English CSW benchmark, CSFleurs \cite{yan2025cs}. 
We exclude evaluating on synthetic test sets generated via LLM and TTS as this yields limited insights. We employ Point-of-Interest Error Rate (PIER) alongside WER to isolate errors on embedded-language (English) words, enabling precise assessment of code-switching-specific performance changes.

\subsection{Code-switching text generation from monolingual prompts}
\label{subsec:csw_text_generation}
To generate synthetic German-English code-switched text, we use GPT-4o with temperature 0.3, prompted with carefully designed linguistic rules that define how English words are morphosyntactically integrated into German matrix sentences. 
The prompt enforces strict single-word substitution following the equivalence constraint theorem \cite{poplack1981formal}: 
the inserted English word must match the original German word in syntactic category, valency, reflexivity, and register. Crucially, German morphology is applied to the English base form, for example, an English verb inserted into a German sentence receives German inflectional endings ("managen", "managst"), and nouns receive German gender, case, and capitalization ("ein Tool", "der Upload"). 
This ensures the resulting utterances reflect the genuine morphological integration patterns observed in naturalistic German-English code-switching, rather than naive word substitution.
The prompt additionally instructs the model to wrap each substituted word in delimiter tags (§§...§§), which we exploit downstream to automatically identify code-switch points for TTS stitching (Section~\ref{subsec:multilingual_tts_stitching}).
This annotation comes at no additional cost, requiring no manual labeling of switch points. 
Full prompt details and linguistic rules are provided in our repository.\footnote{\githubsmall{}\url{https://github.com/enesyugan/robust-code-switching-asr}}

\subsection{Multilingual TTS and stitching}
\label{subsec:multilingual_tts_stitching}
After generating the code-switching text, we use another strong pre-trained multilingual TTS model, namely x-tts-v2 \cite{casanova2024xtts}\footnote{\huggingfacesmall{}https://huggingface.co/coqui/XTTS-v2}.
This model supports many languages and has competitive synthesis quality in all of them. 
The model additionally provides 58 speaker embeddings, enabling substantial speaker diversity in the generated data.
To this end, we tried 3 different approaches.
\textit{German:} We synthesized the whole text, telling the model the transcript is German language. 
\textit{English:} We did the same thing telling the model the text is in English language. 
\textit{Stitching:} The §§...§§ delimiter tags produced during text generation (Section~\ref{subsec:csw_text_generation}) are used to automatically segment each transcript by language, enabling per-segment TTS synthesis without any additional alignment step.
Afterwards, we appended the speech by cutting out zero tails and smoothing the audios at borders. 
Listening to the audios the best results were achieved using the \textit{Stitching} approach, thus, we utilize this version, in our following experiments.

\subsection{Adaptation methods}

Low-Rank Adaptation (LoRA) \cite{pham2021efficient,hu2022lora} is a standard approach for fine-tuning large foundation models, introducing trainable low-rank matrices $\mathbf{A} \in \mathbb{R}^{d_{\text{out}} \times r}$ and $\mathbf{B} \in \mathbb{R}^{r \times d_{\text{in}}}$ as a parallel update path: $\Delta \mathbf{W} = \frac{\alpha}{r}\,\mathbf{A}\mathbf{B}$, with $r \ll \min(d_{\text{in}}, d_{\text{out}})$.
While efficient, LoRA adapters trained on a new data distribution risk overwriting previously learned generalizations when the incoming data distribution at inference time is unknown \cite{ugan2025weight}.

This bottleneck was addressed with the introduction of Bayesian low rank adapters (BLoRA) \cite{ugan2026bayesian}.
The idea is to introduce a Bayesian prior on learned weights $\mathbf{A}$ and $\mathbf{B}$, which changes fixed weight matrices into learned distributions:
\[
q_\phi(A_{ij}) = \mathcal{N}(\mu_{ij}, \sigma_{ij}^2), \quad
q_\phi(B_{ij}) = \mathcal{N}(\mu'_{ij}, {\sigma'_{ij}}^{2}).
\]
The priors for $\mu$ and $\sigma$ are set to $0$ and $0.01$, thus pushing the values of the resulting distribution towards zero, effectively yielding significantly sparser $\Delta \mathbf{W}$ adaptation matrices.
In contrast to dense adapters which manipulate the whole linear layer $\mathbf{W}$, sparse adapters change significantly fewer weights thus having more stable adaptation properties for new subsets of data.

\section{Experiments and Results}
\subsection{Experimental Setup}
\label{sec:experimental_setup}
For fine-tuning we used the same setup for:
\begin{itemize}
    \item \textbf{LoRA}: Standard low-rank adaptation (rank \(r = \) 32)
    \item \textbf{BLoRA}: Bayesian Low-Rank Factorization with KL regularization (rank \(r = \) 32, \(\lambda_{\text{KL}} = \) 0.5 )
\end{itemize}
with learning rate $1e^{-3}$, warmup steps $2000$ and weight decay of $5e^{-4}$ for a maximum of $30000$ steps, used with Whisper v3 turbo.
Following \cite{ugan2026bayesian}, we set $\lambda_{\text{KL}} = 0.5$, which was shown to be a robust default across adaptation tasks.

For code-switching evaluation we utilize the newly published CSFleurs \cite{yan2025cs} dataset. \footnote{We also evaluated on DECM \cite{ugan2024decm}. LoRA degrades substantially, whereas BLoRA remains near baseline. We attribute the residual gap to acoustic mismatch between read-speech training data and DECM's conversational speech.}
Preservation of the models' initial capabilities (backwards testing) is done on the  CommonVoice 14.0 \cite{ardila2020common} (CV) dataset.
We choose CV as it is read speech covering diverse topics, and because our synthetic data was generated from CV transcripts, making it a conservative backward test whose domain closely matches our training distribution, as discussed in Section~\ref{subsec:adapting_first}.
We evaluate our experiments using the standard Word Error Rate (WER) and newly proposed Point-of-Interest Error Rate (PIER) \cite{ugan2025pier}.
While WER focuses on the overall transcription performance, PIER focuses on transcription errors at code-switched words, directly measuring whether our approach improves the model's linguistic switching capacity rather than merely adapting to the general data distribution.

We manually annotated CSFleurs for English code-switch points following a set of explicit annotation guidelines. 
A token is marked as a point of interest if it represents an active insertion of a foreign communicative code into the German syntactic frame, covering lexical insertions, English function words, hybrid morphological forms, and English phrases, while explicitly excluding proper names, internationally standardized codes, and fully integrated German loanwords. 
The annotated file and full annotation guidelines are available in our repository.\footnote{\githubsmall{}\url{https://github.com/enesyugan/robust-code-switching-asr}} All model adaptation experiments were built upon the open-source implementation available at: \footnote{\githubsmall{}\url{https://github.com/enesyugan/continual-asr}}


\subsection{Adapting with filtered data}
\label{subsec:adapting_first}
In our first experiment we evaluate the effect of different amounts of synthetic data on the model's performance. 
To that end we compare commonly used LoRA and newly proposed BLoRA adaptations in Table~\ref{table:amounts_of_data}.
During text synthesis we saw significant hallucinations of the TTS model specifically in cases of short single word texts.
For each segment after synthesizing the text we produce a transcription hypothesis using whisper medium and filter out utterances in which any segment has a CER of $\geq 40\%$.
Additionally, we compare against three variations of a recently suggested model adaptation approach \cite{nguyen2025can}, utilizing a more complex data generation pipeline.
We use the same 10k utterance for data generation as in all our experiments, and use a strong MT system based on DeltaLM \cite{ma2021deltalm}, fine-tuned with German-English translation dataset, described in \cite{mullov2025few}.
Instead of HMM-GMM models we fine-tuned two wav2vec2-xlsr \cite{conneau2020unsupervised}\footnote{\huggingfacesmall{}https://huggingface.co/facebook/wav2vec2-large-xlsr-53} CTC models, for both languages separately and apply forced alignment using these models.
Following \cite{nguyen2025can}, we apply three switching strategies: replacing one word (\textit{1word}), three words (\textit{3word}), or each word with 20\% probability (\textit{0.2}), with splicing and amplitude normalization throughout.

Table~\ref{table:amounts_of_data} reveals a key insight: simply adapting with LoRA adapters does not help a strong baseline model improve code-switching at all.
Although CV and CSFleurs data is read speech we can appreciate substantial WER degradation on all datasets for any amount of data used. 
This suggests that performance in different distributions, such as spontaneous speech, will experience even worse degradation.
Even most recent and multi-stage complex data synthesis approaches show significant degradation not just in WER but even in the transcription of code-switched words, as is shown by the absolute PIER drop of 10.9\% with the best performing augmentation method (3word, row 3).

While degradation reduces with more data, results remain unusable even at 246,503 utterances.
In contrast, when combining synthetic data with BLoRA adaptation, slight overfitting is observed with only 1,000 utterances.
Using more diverse data (10k utterances) already significantly reduces overfitting and actually improves on the CSFleurs code-switching dataset as well.
Using all generated data achieves a relative improvement of \SI{5.31}{\%} WER over an already strong performance, while standard LoRA WER degrades by \SI{74.24}{\%}, in the best case.
The best performing setup improves embedded word transcription by 21.63\% PIER significantly enhancing the already strong model with code-switching capabilities.
\begin{table}[t]
\centering
\scriptsize
\caption{WER $\downarrow$ and PIER $\downarrow$ in \% with Xk utterances. Filter with CER 40\%.}
\begin{tabular}{lccccc}
\toprule
\textbf{\# utterances} & \textbf{Setup} & \multicolumn{1}{c}{\textbf{German}} & \multicolumn{1}{c}{\textbf{English}} & \multicolumn{2}{c}{\textbf{CSFleurs}} \\
 &  & \textbf{WER} & \textbf{WER} & \textbf{WER} & \textbf{PIER} \\
\midrule
 & Whisper & 8.53 & 13.56 & 11.49 & 26.59 \\
\midrule
10k & \cite{nguyen2025can} 1word & 22.54 & 44.05 & 28.81 & 38.91 \\
10k & \cite{nguyen2025can} 3word & 30.78 & 47.9  & 35.51 & 37.49 \\
10k & \cite{nguyen2025can} 0.2   & 26.84 & 50.67 & 33.89 & 43.09 \\
\midrule
1k & BLoRA & 11.59 & 15    & 13.31 & 23.60 \\
1k & LoRA  & 44.15 & 66.05 & 66.00 & 82.30 \\
\hline
10k & BLoRA & 9.77 & 13.68 & 11.37 & 22.25 \\
10k & LoRA  & 20.80 & 50.47 & 33.61 & 62.14 \\
\hline
20k & BLoRA & 9.31 & 13.35 & 11.09 & 21.58 \\
20k & LoRA  & 17.69 & 49.19 & 30.21 & 56.46 \\
\hline
246k & BLoRA & 9.29 & 13.59 & \textbf{10.88} & \textbf{20.84} \\
246k & LoRA  & 13.00 & 33.19 & 20.02 & 43.47 \\
\bottomrule
\end{tabular}
\label{table:amounts_of_data}
\end{table}

The key finding is that for already-strong multilingual models, the data generation pipeline has limited influence on adaptation success. 
Naive fine-tuning on any new data distribution primarily causes the model to lose its generalization capabilities.
Rather than investing in complex synthesis pipelines, the more meaningful focus is on robust knowledge integration: selectively merging switching-relevant linguistic knowledge into the existing model without overwriting it.

\subsection{Speaker vs Text diversity}
We conduct a controlled ablation at a fixed budget of 6,535 utterances to isolate whether robustness gains are driven by linguistic or acoustic coverage. While prior work emphasizes text design and switch patterns \cite{adel2013recurrent, winata2018code}, others show benefits from increasing speaker variability \cite{ogun2025exhaustive}. We contrast a text-rich/speaker-poor setting (TextRich: 6,535 unique transcripts, single speaker) with a speaker-rich/text-poor setting (SpeakerRich: 58 speakers, 117 unique transcripts), combined with BLoRA adaptation.

Table~\ref{table:spk_txt_relatedwork} supports the finding from Section~\ref{subsec:adapting_first} that the most impact is gained from how new data distribution knowledge is integrated to the base knowledge of the model, rather than how the data is generated.
On clean test data, textual diversity yields slightly larger PIER gains than speaker diversity (18.94\% vs. 16.85\%).


\begin{table}[t]
\centering
\scriptsize
\caption{WER $\downarrow$ and PIER $\downarrow$ in \%. TextRich: one speaker with 6535 different texts. SpeakerRich: 58 speakers and 117 different texts.}
\label{table:spk_txt_relatedwork}
\begin{tabular}{lccccc}
\toprule
\textbf{} & \textbf{Setup} & \multicolumn{1}{c}{\textbf{German}} & \multicolumn{1}{c}{\textbf{English}} & \multicolumn{2}{c}{\textbf{CSFleurs}} \\
 &  & \textbf{WER} & \textbf{WER} & \textbf{WER} & \textbf{PIER} \\
\midrule
 & Whisper & 8.53 & 13.56 & 11.49 & 26.59 \\
\midrule
TextRich & BLoRA & 9.15 & \textbf{13.16} & \textbf{10.85} & \textbf{21.58} \\
SpeakerRich & BLoRA & 10.33 & 14.11 & 12.03 & 22.11 \\
\bottomrule
\end{tabular}
\vspace{-2em}
\end{table}

\subsection{Filtering Synthetic Data}
\label{subsec:diff_filters}
A practical question for any synthetic data pipeline is how much filtering is worth the effort. During synthesis we observed significant TTS hallucinations, particularly for short, single word segments, but it is unclear whether filtering these improves adaptation or is simply unnecessary overhead.
To answer this question we train four variants using BLoRA, filtering synthetic sub-segments by CER thresholds of 5\%, 20\%, and 40\% (computed via Whisper-medium), plus an unfiltered baseline using all 10k transcripts across 58 speakers.
All models are trained with BLoRA adapters as described in Section~\ref{sec:experimental_setup}.

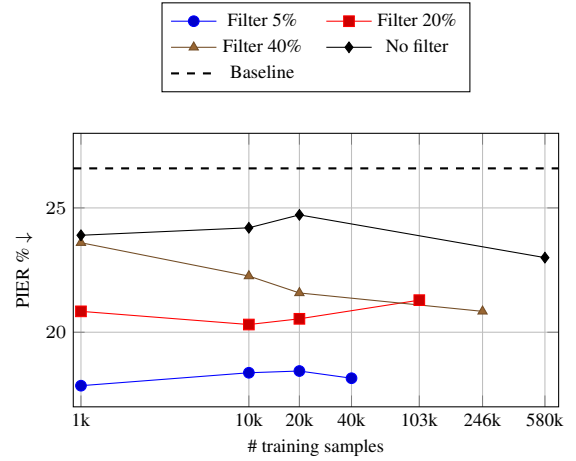
\begin{figure}[t]
\centering
\scriptsize
\begin{tikzpicture}
\begin{axis}[
    width=\linewidth,
    height=0.65\linewidth,
    xlabel={\# training samples},
    ylabel={PIER \% $\downarrow$},
    xmode=log,
    log basis x=10,
    xmin=900, xmax=700000,
    ymin=17, ymax=28,
    grid=both,
    legend style={
        at={(0.5,1.15)},
        anchor=south,
        legend columns=2
        row sep=2pt,
        /tikz/every even column/.append style={column sep=1em}
    },
    xtick={1000,10000,20000,40869,103550,246503,580000},
    xticklabels={1k,10k,20k, 40k,103k,246k,580k},
]

\addplot+[mark=*] coordinates {
    (1000   , 17.85)
    (10000  , 18.37)
    (20000  , 18.44)
    (40869  , 18.15)
};
\addlegendentry{Filter 5\%}

\addplot+[mark=square*] coordinates {
    (1000   , 20.84)
    (10000  , 20.31)
    (20000  , 20.54)
    (103550 , 21.29)
};
\addlegendentry{Filter 20\%}

\addplot+[mark=triangle*] coordinates {
    (1000   , 23.60)
    (10000  , 22.26)
    (20000  , 21.58)
    (246503 , 20.84)
};
\addlegendentry{Filter 40\%}

\addplot+[mark=diamond*] coordinates {
    (1000   , 23.90)
    (10000  , 24.20)
    (20000  , 24.72)
    (580000 , 23.00)
};
\addlegendentry{No filter}

\addplot[
    black,
    dashed,
    thick,
] coordinates {
    (900, 26.59)
    (700000, 26.59)
};
\addlegendentry{Baseline}

\end{axis}
\end{tikzpicture}
\caption{PIER $\downarrow$ on CSFleurs using different amounts of data and different CER quality filters during text synthesis.}
\label{fig:filter_pier}
\vspace{-2em}
\end{figure}

We report PIER scores of our models trained with different filtering strategies and different amounts of data in Figure~\ref{fig:filter_pier}.
In this figure, we can see that filtering for quality is especially important when only small amounts of training data are available.
For only 1000 training samples a filter of 40\% does not make a difference to no filtering at all, however, a clear trend emerges: with more data, the negative effect of hallucinations diminishes.
Most strikingly, with 17.85\% PIER, the best score is achieved by the most aggressive filter of 5\% CER, reducing code-switching related errors by 32.87\% already with as few as 1000 samples.

To complement the quantitative analysis, we present a qualitative example illustrating 
the type of error the adapted model learns to overcome.
In the following example, the English word \emph{matter} is repeatedly overridden by the base model in favour of the German token \emph{meta}, which is acoustically close (\textipa{/\textprimstress m\ae t\textschwa r/} vs.\ \textipa{/\textprimstress me\text{:}ta/}) and substantially more frequent in German text. The base model resolves the ambiguity towards the more probable German token, ignoring the English code-switch. 
\begin{tcolorbox}[
  colback=gray!3,
  colframe=black!40,
  title=Qualitative Example: Language Model Prior Bias in Code-Switching,
  fonttitle=\bfseries,
  boxrule=0.5pt
]
\scriptsize
\textbf{Reference (German + English):} \\
\emph{alles im universum besteht aus matter sämtliche matter besteht aus tiny particles die atoms genannt werden}

\vspace{0.5em}
\textbf{Base Whisper model:} \\
\emph{alles im universum besteht aus \textbf{meta} sämtliche   \textbf{meta} besteht aus tiny particles die atoms genannt werden}

\vspace{0.5em}
\textbf{BLoRA (5\% CER filter, 1k samples):} \\
\emph{alles im universum besteht aus \textbf{matter} sämtliche \textbf{matter} besteht aus tiny particles die atoms genannt werden}
\end{tcolorbox}

\section{Conclusion}
Code-switching ASR adaptation is often framed as a data problem: synthesize better code-switching speech, and performance will follow. 
This paper demonstrates that this framing breaks down precisely where it matters most: 
when the base model is already strong. 
We show that standard fine-tuning on synthetic data consistently degrades both monolingual and code-switching performance regardless of data quantity or pipeline complexity, and that even state-of-the-art multi-step generation approaches fail in this setting. 
The root cause is integration quality, not data quality: 
naive adaptation overwrites existing linguistic knowledge rather than extending it. 
BLoRA addresses this with sparse, uncertainty-regularized adaptation matrices that selectively incorporate switching-relevant knowledge while preserving monolingual representations, achieving meaningful WER and PIER improvements on CSFleurs while maintaining CommonVoice performance.

The broader lesson is one that extends beyond code-switching: 
as the community increasingly builds on strong foundation models, adaptation methodology matters more than data methodology. 
Adding a new capability to an already powerful model is fundamentally different from training that capability from scratch, and should be treated as such.
We hope this work encourages future research to evaluate adaptation approaches on strong baselines, and to treat knowledge integration, not data complexity, as the primary design variable.

\section{Generative AI Use Disclosure}
The authors used generative AI tools only for language editing, readability improvements, and figure editing. Generative AI was not used to generate scientific content, experimental results, data analyses, or conclusions.

\ifcameraready
     
\section{Acknowledgment}
This work was supported in part by the European Union’s Horizon research 
programme under grant agreement No. 101135798,
project Meetween (My Personal AI Mediator for
Virtual MEETtings BetWEEN People) and European Union’s Horizon Europe programme grant
agreement No. 101213369 (DVPS). The authors
gratefully acknowledge computing time provided
on HoreKa at the National High-Performance Computing Center at KIT (NHR@KIT), supported by
the Federal Ministry of Education and Research,
the Ministry of Science, Research and the Arts of
Baden-Württemberg, and the DFG, and Interactive-AI
\else
     
\fi
\bibliographystyle{IEEEtran}
\bibliography{mybib}

\end{document}